# Comparative Analysis of Efficient Adapter-Based Fine-Tuning of State-of-the-Art Transformer Models

Saad M. Siddiqui
ssiddiqui60@gatech.edu

Mohammad Ali Sheikh
msheikh36@gatech.edu

Muhammad Aleem
maleem3@gatech.edu

Kajol Ramesh Singh
ksingh362@gatech.edu

## Abstract

*In this work, we investigate the efficacy of various adapter architectures on supervised binary classification tasks from the SuperGLUE benchmark as well as a supervised multi-class news category classification task from Kaggle. Specifically, we compare classification performance and time complexity of three transformer models, namely DistilBERT, ELECTRA, and BART, using conventional fine-tuning as well as nine state-of-the-art (SoTA) adapter architectures. Our analysis reveals performance differences across adapter architectures, highlighting their ability to achieve comparable or better performance relative to fine-tuning at a fraction of the training time. Similar results are observed on the new classification task, further supporting our findings and demonstrating adapters as efficient and flexible alternatives to fine-tuning. This study provides valuable insights and guidelines for selecting and implementing adapters in diverse natural language processing (NLP) applications.*

## 1. Introduction

### 1.1. Problem and Existing Approaches

Transformer-based [1] models represent the SoTA in terms of NLP applications, having consistently demonstrated unparalleled performance across a wide range of tasks. These models are characterized by deep neural network architectures with large number of learnable parameters, often approaching the order of hundreds of millions, which are optimized through a combination of supervised, self-supervised, and semi-supervised pre-training on vast corpora.

NLP researchers and practitioners are interested in leveraging the capabilities of transformer models for domain-specific tasks, which invariably require updating or tuning learnable parameters. A popular method for doing so is fine-tuning [2], which involves updating either all learnable parameters or a subset of learnable parameters with methods such as gradient descent using loss from a smaller, task-specific labeled dataset. Another popular method is domain-adaptation [3], which involves aligning model performance with a target domain through adversarial training, domain-invariant representations, and other techniques which involve a mix of source and target domain data.

Despite their efficacy, both methods are often expensive in terms of time, often taking prohibitively long owing to the large number of learnable parameters that need to be updated. There is also an additional risk of overfitting and issues with gradient flow, as the absence of appropriate regularization techniques and architectural constructs for gradient flow augmentation can cause problems with generalization when tuning a large number of parameters using small, labeled corpora.

### 1.2. Adapters

Adapters [4] are lightweight task-specific modules, typically consisting of projection layers or small neural networks, that can be inserted between layers in transformer models. During training, only adapter parameters are updated while the rest of the model parameters remain unchanged. This makes adapters a parameter-efficient alternative to finetuning and domain adaptation: adapters allow the retention of generalizable knowledge learned from large corpora as encoded by original model weights while still allowing the parameters of the adapter to learn task-specific information. Contrary to fine-tuning a classification head in conventional transfer learning, the insertion of adapter modules is not limited to later layers in the network.

Successful integration of adapters in transformer-based models can significantly reduce computational resources and time needed for modifying models for a specific domain or task, even with limited labeled data [5]. This parameter efficiency means that adapting models to new tasks will become faster, cheaper, and more accessible, particularly for researchers and practitioners with limited resources. It will enable more widespread use of powerful language models in various applications, fostering innovation and advancing the field of NLP by making state-of-the-art tools available to a broader audience.

### 1.3. Objectives

This project aims to empirically validate and quantify the classification and time complexity benefits of adapters as parameter-efficient alternatives to fine-tuning transformer models for supervised classification. Additionally, the project also aims to identify which adapter designs perform best across a range of transformer models and tasks to understand their advantages over traditional methods.

### 1.4. Datasets

We have used two main datasets in this project.

1. **SuperGLUE [6]**: A collection of diverse and challenging NLP tasks designed to evaluate the performance of language models on understanding and reasoning. It consists of tasks such as question answering, natural language inference, and co-reference resolution, providing a comprehensive assessment of model capabilities.

2. **News Category Classification [7]**: A Kaggle dataset consisting of 210,294 records of news articles from the Huffington Post collected between 2012 and 2022 with predictors such as headlines, authors, and publication date mapped to 42 highly imbalanced categories such as Politics, Wellness, Entertainment, to name a few.

In the SuperGLUE benchmark, we specifically focus on the following binary classification tasks, partly due to their simplicity and partly due to time and resource constraints.

Table 1: Summary of SuperGLUE Tasks

| Task | Description | Training Set Size | Validation Set Size |
|---|---|---|---|
| Boolean Questions | Answering yes/no questions | 9427 | 3270 |
| Commitment Bank | Textual entailment task | 250 | 56 |
| Recognizing Textual Entailment | Binary entailment classification task | 2500 | 278 |

To supplement these tasks, we explored model performance on the Corpus of Linguistic Acceptability (CoLA), Winograd NLI (WNLI), and Microsoft Research Paraphrase Corpus (MRPC) tasks from the GLUE [8] benchmark. **In the interest of brevity, however, we have chosen to focus the rest of the paper exclusively on the tasks in Table 1 and News Classification,** with results for these tasks provided in Appendix II. Exploring the performance of adapters on the News Classification task was motivated by a choice to explore if model performance trends on a binary classification academic benchmark such as SuperGLUE also generalized to a complex, multi-class classification task using real-world data.

## 2. Approach

At a high level, our approach to the problem involved running experiments to benchmark model performance in terms of validation set accuracy as well as training time with and without one each of nine adapters across three SuperGLUE tasks as well as the News Classification task using base variants of three SoTA transformer models, namely DistilBERT [9], ELECTRA [10], and BART [11].

### 2.1. Addressing the Problem

To address our problem, we used HuggingFace's Transformers [12] library along with the Adapters library [13] from Adapter Hub. The Adapters repository provided code for running simple sequential bottleneck adapters on GLUE tasks, but we needed to extend this code to fit our requirements. Specifically, we added support for SuperGLUE, incorporating tasks from Table 1 and extended the repository to include various adapter architectures such as sequential bottleneck, stacked sequential bottleneck Mix and Match, iA3, LoRA, Prefix Tuning, Compacter++, Prompt Tuning, and UniPELT [14]. Additionally, we modified the code to return epoch-level estimates of training and validation loss as well as task-specific model evaluation metrics such as accuracy, F1 score, and Matthew's Correlation. Lastly, we added support to handle traditional fine-tuning on the same tasks for comparison. Our approach therefore resulted in a single pipeline that could be used to benchmark model performance in terms of evaluation metrics and time complexity with and without adapters across all three transformer models of interest.

### 2.2. Novelty

Firstly, our approach is novel in its comprehensive evaluation of a wide range of adapter architectures on state-of-the-art models and complex benchmarks. By including a practical news classification task, we aimed to provide real-world insights into the effectiveness of adapters, highlighting their potential for efficient and flexible adaptation of pre-trained language models to new tasks.

Additionally, by considering epoch-level learning curves in conjunction with end-of-training point estimates of model performance with and without adapters, the comprehensive nature of our analysis provides clear insights into the efficiency and effectiveness of adapters, demonstrating their potential as a viable alternative to full model fine-tuning in NLP tasks. By evaluating a wide range of adapter architectures, we contributed new knowledge to the field, offering practical guidelines for selecting and implementing adapters in various NLP applications.

### 2.3. Hypotheses

We hypothesize that adapters will offer a more efficient and flexible way to modify all three transformer models for both datasets. Specifically, we expect most, if not all, adapters to have comparable performance to full fine-tuning while requiring significantly less training time. We hypothesize that our experiments will yield adapters with varying levels of performance, with more advanced architectures outperforming simpler ones. As such, we also hypothesize that we will successfully identify "reasonable default" adapter configurations which perform well across all tasks and models while minimizing training time.

## 3. Experiments

### 3.1. Setup and Hyperparameters

We began our investigation with SuperGLUE binary classification tasks and set up a grid of experiments which were delegated across team members. Specifically, we added each of 9 adapter configurations to each of 3 transformer models and trained the resulting architecture for 10 epochs, using HuggingFace and AdapterHub defaults for the hyperparameter including a learning rate of 5e-5, batch size of 8, and an Adam optimizer with $\beta_1 = 0.9$, $\beta_2 = 0.999$, and $\varepsilon = 1e-08$. This allowed us to isolate differences in performance exclusively to the presence or absence of adapter configurations while also ensuring each architecture had a good probability of convergence regardless of architecture. Custom code was written to cater to bespoke data preprocessing, tokenization, and evaluation requirements for each SuperGLUE task as well as the integration of a range of adapters across all 3 models.

We anticipated some tasks would take very long to train due to the large size of the datasets and models, as well as the former's inherent preprocessing complexity. Consequently, we used an NVIDIA L4 GPU on Google CoLab and NVIDIA A100 GPU on PACE-ICE to accelerate training. Our initial experiments with a simple sequential bottleneck adapter on the BoolQ task yielded significantly subpar results, which we anticipated due to the relative simplicity of the adapter architecture. This incentivized us to expand our initial search space of adapters to include more advanced architectures such as LoRA, iA3 and Mix-and-Match.

### 3.2. Model Selection

We used three base models: DistilBERT, BART, and Electra base. These models were chosen for their relatively lightweight architecture and the fact that they have not been extensively tested with adapters in existing literature. DistilBERT, with approximately 66 million parameters, is a distilled version of BERT, making it more efficient while retaining a substantial portion of BERT's performance capabilities. BART, with around 140 million parameters, is a transformer model that combines bidirectional and autoregressive transformers, suitable for a wide range of NLP tasks including classification. ELECTRA, consisting of about 110 million parameters, is known for its efficiency and performance in pre-training tasks. These models are well-suited for NLP-based binary classification tasks due to their architecture, which effectively captures contextual information necessary for making accurate predictions.

### 3.3. Adapter Methods

We experimented with various adapter methods to understand how they modify the architecture of each model and their impact on performance. The adapters used were seq bn (Sequential Bottleneck), prefix tuning, MAM (Mix and Match) adapter, compacter plusplus (Compacter++), UniPELT (Unified Prompt and Embedding Learning Techniques), LoRA (LowRank Adaptation), stacked seq bn (stacked sequential bottleneck with tanh and ReLU activations), IA3 (Improved Adapter Architecture with Activation and Attention) and prompt tuning. These adapters were selected based on their innovative approaches to modifying the model architecture, aiming to enhance performance with minimal additional parameters.

We deliberately experimented with a wide range of adapters because of the diversity of architectural modifications and preference biases they exhibited. For instance, adapters like Sequence Bottleneck and its stacked variant introduce bottleneck layers that reduce computational overhead while capturing essential information. As such, these adapters were expected to be more generic in their preference biases and utility across SuperGLUE tasks. In contrast, prefix tuning and prompt tuning focus on optimizing input prompts rather than model weights, and were expected not to perform well as generic, out-of-the-box adapters for classification tasks. MAM and Compacter++ utilize advanced parameter sharing and low-rank decomposition techniques to improve efficiency. UniPELT integrates prompt tuning with embedding learning, while LoRA and iA3 introduce low-rank approximations and improved activation functions, respectively, to boost performance.

### 3.4. Metrics

To compare the performance of fine-tuning and adapter training, we used three primary metrics: time to train, evaluation dataset accuracy and training and validation loss by epoch. These metrics were chosen to provide a clear comparison of the computational efficiency and predictive performance of each method. Time to train is crucial for understanding the resource requirements of each approach, while evaluation accuracy directly reflects the effectiveness of the models and adapters in making accurate predictions. Additionally, training and validation loss by epoch helps monitor the learning process and convergence behavior, providing insights into the stability and efficiency of each training method. Other metrics such as floating point operations were also observed and reported on, but the primary focus of our analysis revolves around the three metrics mentioned previously.

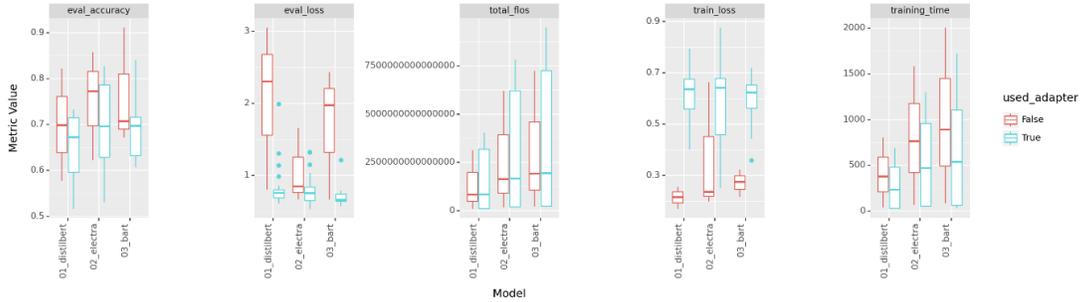

Figure 1: Distributions of Training and Validation Metrics by Model – SuperGLUE

## 3.5. News Classification

All three models were first fine-tuned on the News Classification dataset. Owing to time and resource constraints, we chose to focus on only one of the three models, specifically ELECTRA, for adapter tuning experimentation with the Mix-and-Match and Sequential Bottleneck (ReLU) adapters, partly because these adapters had demonstrated good performance on concurrent SuperGLUE experiments and partly because their architecture seemed to be intuitively compatible with the requirements of the News Classification problem. The data was converted to the Hugging Face Dataset format, and a tokenization function was applied to the text headlines, truncating and padding them to a maximum length of 128 tokens. This prepared the data for efficient training with transformer models. We also implemented custom preprocessing and metrics specific to the news classification task and analyzed the accuracy, F1-score, and loss curves for each configuration, in addition to measuring training time.

## 4. Results and Discussion

### 4.1. SuperGLUE Performance – Results

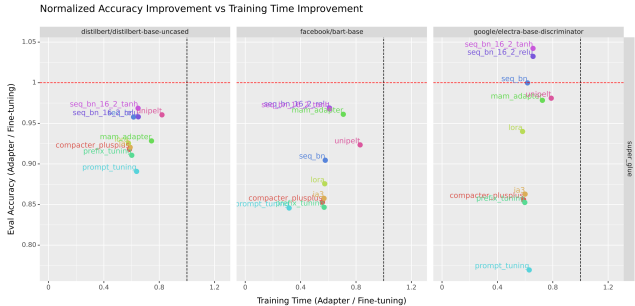

Figure 2: Normalized Adapter Accuracy and Train Time

Table 2 summarizes the mean validation accuracy and training time across all SuperGLUE tasks with each model and adapter configuration, including models trained with full fine-tuning. This is complemented by Figure 1, which shows the distribution of additional metrics such as training and validation binary cross-entropy loss as well. Figure 2 compares the normalized accuracy and training time of adapters each adapter relative to fine-tuning across models.

Table 2: Average SuperGLUE Performance

| Adapter | Average Validation Set Accuracy | | | Average Training Time (seconds) | | |
|---|---|---|---|---|---|---|
| | Distil BERT | BART | ELECTRA | Distil BERT | BART | ELECTRA |
| No Adapter | 0.699 | 0.763 | 0.750 | 405.10 | 993.36 | 804.64 |
| Compacter ++ | 0.642 | 0.651 | 0.642 | 237.04 | 553.71 | 474.07 |
| IA3 | 0.644 | 0.655 | 0.648 | 238.93 | 565.35 | 483.08 |
| LoRA | 0.647 | 0.668 | 0.705 | 234.21 | 570.38 | 468.38 |
| MAM | 0.649 | 0.734 | 0.734 | 301.29 | 703.61 | 583.44 |
| Prefix Tuning | 0.637 | 0.646 | 0.640 | 243.64 | 566.23 | 481.23 |
| Prompt Tuning | 0.623 | 0.646 | 0.577 | 257.72 | 315.34 | 506.99 |
| Sequential Bottleneck | 0.670 | 0.691 | 0.750 | 248.97 | 574.49 | 497.05 |
| Bottleneck (ReLU) | 0.670 | 0.740 | 0.775 | 262.67 | 605.00 | 529.23 |
| Bottleneck (tanh) | 0.677 | 0.739 | 0.782 | 262.09 | 605.34 | 529.14 |
| UniPELT | 0.672 | 0.705 | 0.736 | 331.83 | 824.21 | 634.89 |

### 4.2. SuperGLUE Performance – Discussion

**Sequential Bottleneck Adapters**

Results from Table 2 and Figures 1 – 2 validate many of our initial hypotheses. Firstly, they demonstrate that adapters consistently lead to performance comparable to or, in some cases, even better than traditional fine-tuning in terms of evaluation accuracy while requiring significantly less training time. This trend holds across all three models investigated in the experiments. In particular, the Sequential Bottleneck adapter and its variants tend to achieve >= 90% of the validation accuracy of full fine-tuning with ELECTRA despite requiring, <= 70% of the training time. The improvement in training time is true for all adapters, although different adapters achieve varying degrees of success in terms of validation set accuracy. This is attributable to their architectural biases and how they lend themselves to binary classification. For instance, the superior performance of sequential bottleneck adapters can be attributed to their two sequential bottleneck layers, allowing the model to capture more intricate patterns.

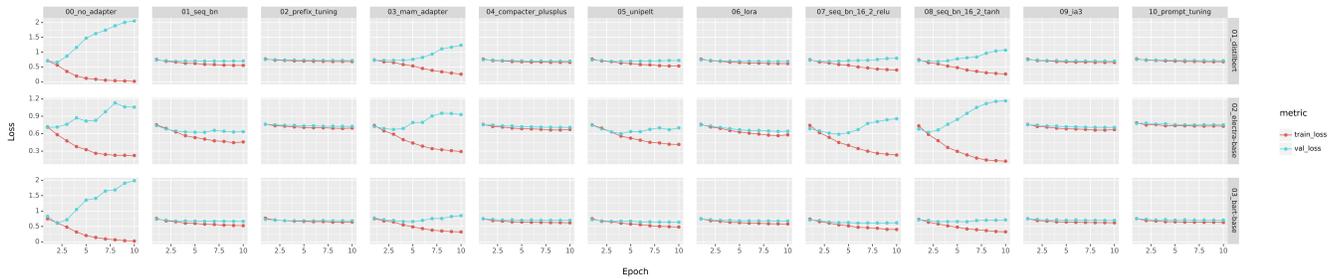

Figure 3: Average epoch-level Training and Validation Loss by Adapter and Model across all 3 SuperGLUE tasks

**Prompt Tuning and Prefix Tuning**

In contrast, Prompt tuning and Prefix tuning adapters perform poorly across tasks due to their task-specific nature, consistently leading to a very low accuracy relative to fine-tuning despite little evidence of overfitting. These adapters are less effective for classification tasks because their design suits tasks needing rich contextual understanding and generation rather than straightforward classification. Prefix Tuning prepends trainable prefix tokens to the input, modifying keys and values in the multi-head attention mechanism. While this enhances sequence-to-sequence tasks like text generation, it does not benefit classification tasks, which rely on fixed input representation and feed-forward layers to capture discriminative features. Likewise, Prompt Tuning adds a fixed prompt to the input, which is effective for generative tasks but less so for classification, which needs a clear decision boundary and efficient feature transformation.

**Mix-and-Match**

Conversely, the Mix and Match (MAM) adapter performs well because it balances simplicity and complexity by combining Prefix Tuning and parallel bottleneck adapters. This architecture allows it to dynamically adapt to different task requirements, making it suitable for classification. Parallel Bottleneck Adapters are lightweight modules within the transformer's feed-forward blocks, efficiently handling complex transformations through a combination of down-projection, non-linearities, and up-projection layers, helping them extract relevant features from the input data.

**Overfitting with Fine-Tuning**

Training and validation set loss for most adapters generally does not show a large divergence, as is evident from Figures 1 and 3. However, fine-tuning almost always results in training loss converging to 0 while validation set loss diverges. This validates our earlier hypotheses about fine-tuning of large transformer models having a higher likelihood of overfitting compared to adapters. Interestingly, validation set accuracy with fine-tuning remains at or above adapter performance despite poor validation loss. This suggests fine-tuning results in models that become increasingly confident about wrong predictions in probabilistic inference space, but do not necessarily make more or fewer wrong predictions [15].

**Floating Point Operations**

The addition of adapters to a model does not result in a substantial increase in its median floating-point operations. However, as shown in Figure 1, the distribution of operations with adapters is generally skewed towards a higher number of operations. This intuitively makes sense as the inclusion of submodules with varying degrees of complexity within existing architectures should ideally increase the number of operations performed during forward passes. This suggests that while adapters do make substantial improvements in training time, they do so at a potential cost of incurring more computations.

### 4.3. News Classification Dataset - Results and Discussion

Table 3: ELECTRA Results – News Classification

| Adapter | Training Time | Validation Loss | Validation Accuracy | Validation F1 Score |
|---|---|---|---|---|
| None | 2h 33min | 1.325 | 0.637 | 0.625 |
| MAM | 2h 02min | 1.350 | 0.614 | 0.596 |
| Seq.Bottleneck (ReLU) | 1h 05min | 1.575 | 0.567 | 0.532 |

Based on promising results from the SuperGLUE experiments, we selected the ELECTRA model and the MAM and Sequential Bottleneck (ReLU) adapters for experiments with the News Classification dataset, as discussed in 3.5. Table 3 shows the performance of this set of experiments, with learning curves shown in Figure 4 - 6.

The fine-tuned model (without adapters) achieved the highest evaluation accuracy (0.637) and F1 score (0.625), indicating that full fine-tuning can still be the most effective method for achieving high performance on the news classification task and also had the lowest evaluation loss (1.325), which aligns with its higher accuracy and F1 score. The MAM adapter showed a slight drop in performance with an evaluation accuracy of 0.614 and F1 score of 0.596. It also had slightly higher validation set loss. Sequential bottleneck (ReLU) had the lowest accuracy (0.567), F1 score (0.532) among the three configurations and higher validation set loss (1.575). Training time was significantly reduced with adapters. The fine-tuned model took 2 hours and 33 minutes, whereas the MAM adapter reduced the time to

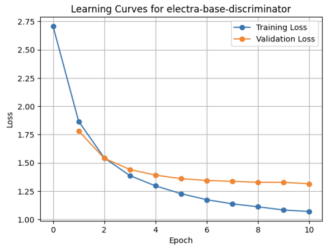
Figure 4: Fine-tuning
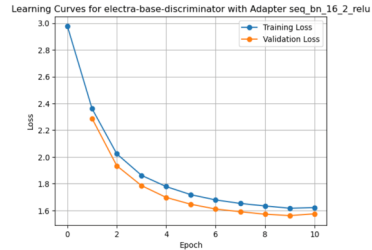
Figure 5: Seq_Bn_ReLU
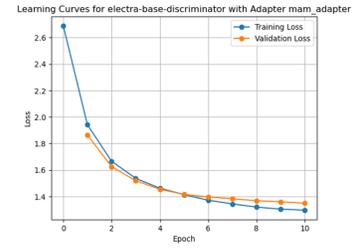
Figure 6: Mix-and-Match

2 hours and 02 minutes, and the Sequential Bottleneck adapter further reduced it to 1 hour and 55 minutes. This demonstrates one of the primary advantages of using adapters: reduced training time. Learning curves suggest that without adapters, the model does learn effectively but shows signs of overfitting, as indicated by the gap between training and validation losses. With the addition of the MAM adapter, the model shows smooth convergence and less overfitting, suggesting that the adapter provides beneficial regularization. In contrast, with Sequential Bottleneck (ReLU) adapter, the model learns effectively but still shows signs of overfitting, although to a lesser extent compared to the model without adapters. These results suggests that general observations about incremental benefit of adapters in terms of training time does indeed far outweigh the relative drops in performance even on a multi-class supervised classification task, with some adapters showing weak regularizing effects.

## 5. Conclusions

The results from our experiments demonstrate that adapter-based training provides an effective and efficient alternative to traditional fine-tuning for NLP classification tasks. Adapters generally achieve comparable performance to full fine-tuning while significantly reducing training time in both binary and multi-class classification contexts.

Specific adapters, such as Mix-and-Match (MAM), exhibit robust performance on SuperGLUE, balancing modeling complexity parameter efficiency. Sequential bottleneck adapters with tanh and ReLU activations show a tendency perform well on more complex SuperGLUE tasks despite some evidence of minor overfitting, suggesting that their stacked architecture is better suited for handling more intricate datasets. Conversely, adapters like LoRA and Sequential Bottleneck adapters without ReLU and tanh activations tend to underfit on complex tasks, indicating a need for further refinement in their design for such applications. Furthermore, adapters like Prompt Tuning and Prefix Tuning, while powerful for generative tasks, do not align well with the requirements of classification due to their architectural focus on contextual understanding and sequence generation rather than feature extraction and decision-making. As such, we successfully verified our hypothesis of adapters yielding variable performance depending on their architectures. Most importantly, however, the benefit of adapters seems to be consistent across multiple transformer-based models, thereby establishing them as a generally viable alternative to fine-tuning across multiple contexts. However, due to time and resource constraints, we were unable to explore if the same trends held on other tasks within the SuperGLUE benchmark, which is an area for future work.

With regards to the News Classification task, our results demonstrate that adapter-based tuning might not always outperform traditional fine-tuning, highlighting the importance of task-specific considerations and suggest the need for further refinement and task-specific tuning. Our findings support the continued exploration of adapter-based methods, with a focus on optimizing configurations and addressing dataset-specific challenges to maximize their effectiveness.

## 6. Future Work

Future experiments could expand on these findings by exploring larger transformer models with more parameters, such as BERT-large or GPT-3, to evaluate the scalability and effectiveness of adapters. Additional hyperparameter tuning could provide deeper insights into optimizing adapter performance across various tasks. However, scaling for models with millions of parameters presents significant computational challenges. Solutions like NVIDIA TensorRT could be employed to optimize inference performance and reduce the computational load. Moreover, the difficulties associated with smaller datasets such as CB underscore the need for high-quality, well-annotated datasets to achieve reliable performance evaluations. Future work could involve curating or identifying more robust datasets with higher evidence and clearer task definitions to mitigate these challenges. Additionally, leveraging advanced data augmentation techniques and synthetic data generation could help improve the robustness and generalizability of models trained with adapters.

In conclusion, while adapter-based training offers significant advantages in terms of efficiency and performance, future research should focus on scaling these methods to larger models, optimizing hyperparameters, and addressing dataset-related challenges to fully realize their potential in NLP classification tasks. These efforts will not only enhance model performance but also contribute to the development of more efficient and scalable solutions for a wide range of NLP applications.

# Appendix I

## A. Project Code

All code for our project, including experimental notebooks, intermediate artifacts, and data, are stored in our project's GATech GitHub repository, which can be found at: https://github.gatech.edu/ssiddiqui60/CS7643-final-project

The SuperGLUE experiment code used a combination of the HuggingFace Transformers and Adapters libraries with example code from HuggingFace and Adapter-Hub documentation. We specifically modified the code to

1. Support SuperGLUE instead of the default GLUE
2. Add a mapping between SuperGLUE tasks and their corresponding DatasetDict I/Os
3. Made the script generalizable enough to run for any binary classification task with any supported transformer model on the HuggingFace Hub and any Adapter supported through Adapters Library
4. Log training and validation set loss as well as task-specific metrics such as accuracy, F1 score, and Matthew's correlation at an epoch level as well.
5. Generate learning curves of validation and training loss by epoch
6. Export epoch level and end-of-training point estimates of time complexity and model evaluation metrics to a centralized repository for logging
7. Generating all visualizations used in the report as well as supplementary visualizations by task and model for extracting model insights.

For the News Classification task, we used HuggingFace and Adapter Hub documentation to write a custom script from scratch to load data, tokenize it for each model, and run similar experiments as for SuperGLUE. Code for plotting and visualizing loss curves was developed as well along with code for logging epoch-level metrics.

## B. Work Division

Delegation of tasks among team members was crucial for successful execution of the project. Muhammad Ali Sheikh collected results for the SuperGLUE RTE and CB tasks along with an auxiliary SuperGLUE ReCORD task that we chose not to include in the main report due to limited space. Ali also worked Muhammad Aleem initially found the base code for GLUE tasks using a single adapter and extended it to include SuperGLUE and other adapters. Saad M. Siddiqui modified the base code for data collection, visualization, and aggregation which informed most of the results and conclusions presented in the report. In addition to collecting data for the BoolQ task, Saad also contributed to the report. Kajol R Singh focused on the news classification task, reusing the existing base code and making necessary modifications to adapt it for the news classification dataset, and conducting experiments with all three models, including their adapter configurations.

| Student Name | Contributed Aspects | Details |
|---|---|---|
| Mohammad Ali Sheikh | RTE, CB and ReCORD experiments. Debugging errors in SuperGLUE and News Classification code. Instructing team in use of PACE-ICE | Ali used code developed for SuperGLUE experiment notebooks to collect results for CB and RTE tasks from the SuperGLUE benchmark. He also developed a custom version of the notebook for the ReCORD task. Ali also helped all team members configure and use PACE-ICE for their experiments. |
| Muhammad Aleem | Wrote base script, contributed extensively to SuperGLUE section of model report | Implemented the base script to run base experiments on SuperGLUE tasks with the 9 adapters mentioned in the paper. Aleem also helped the team explain results collected during experiments with his knowledge of deep learning and machine learning, and wrote the first draft of the report. |
| Saad M. Siddiqui | BoolQ experiments, custom code for logging and visualization, report organization and content, repository maintenance, auxiliary GLUE experiments | Collected results for BoolQ task. Also modified base script with code for non-loss metrics, epoch-level metric logging, storing and exporting experiment results, generating validation curves, generating all SuperGLUE visualizations, and result aggregation. Contributed to all sections of the report. Maintained project repository on GitHub. Also collected results for auxiliary GLUE MRPC, COLA, and WNLI tasks. |
| Kajol Ramesh Singh | News Category Classification Task – implementation, experiments and analysis | Made necessary modifications to adapt base code for the news category dataset, and ran experiments for all three models, with fine tuning and with multiple adapters. See new_dataset*.ipynb notebooks. Kajol also wrote the content for the sections of the report related to the News Classification Task. |

## C. References

[1] Merritt, Rick. "What Is a Transformer Model?" *NVIDIA Blog*, 25 Mar. 2022, blogs.nvidia.com/blog/what-is-a-transformer-model/.

[2] Gugger, Sylvain. "Fine-Tune a Pretrained Model." *Huggingface.co*, 14 June 2021, huggingface.co/docs/transformers/en/training.

[3] Meer, Anne-Maj van der. "Domain Adaptation: Types and Methods." *Www.taus.net*, 19 Dec. 2022, www.taus.net/resources/blog/domain-adaptation-types-and-methods. Accessed 29 July 2024.

# Appendix II – Additional Results

## A. Average Validation Set Accuracy by Adapter – GLUE vs SuperGLUE

As mentioned in Section 1.4, we also explored model performance with and without adapters on binary classification tasks from the GLUE benchmark. We found the SuperGLUE tasks to be more discriminative in terms of model performance and training time, which is why we chose to focus our analyses on the SuperGLUE tasks. Our results from the GLUE experiments are included for completeness.

| | Validation Accuracy | |
|---|---|---|
| **Adapter** | **GLUE** | **SuperGLUE** |
| **mam_adapter** | 0.6430 | 0.7057 |
| **seq_bn** | 0.6019 | 0.7035 |
| **lora** | 0.6014 | 0.6737 |
| **unipelt** | 0.5921 | 0.7042 |
| **seq_bn_16_2_relu** | 0.5827 | 0.7281 |
| **seq_bn_16_2_tanh** | 0.5733 | 0.7327 |
| **ia3** | 0.5610 | 0.6486 |
| **compacter_plusplus** | 0.5577 | 0.6451 |
| **prompt_tuning** | 0.5481 | 0.6153 |
| **prefix_tuning** | 0.5467 | 0.6410 |

## B. Time, Loss, and Accuracy Distributions by Benchmark

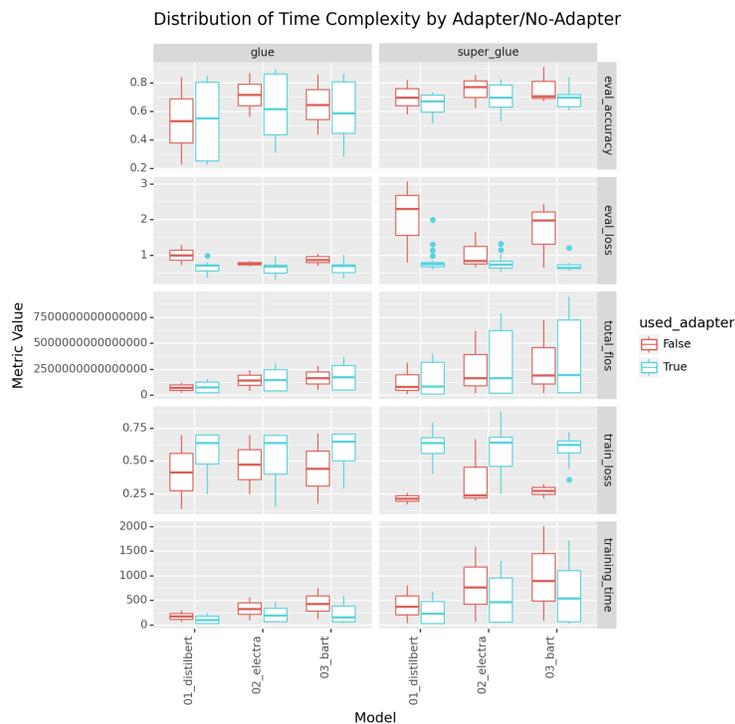

## C. Learning Curves by Task – SuperGLUE

The learning curves presented in Figure 3 and discussed in Section 4.2 represent the mean epoch-level cross-entropy loss with and without adapters across all tasks in the SuperGLUE benchmark. We used the mean for the visualization to conserve space and condense trends to fewer plots, as most task-level training and validation losses demonstrated the same trend as the mean loss. For completeness, we provide the task-level learning curves for both SuperGLUE and GLUE benchmarks here.

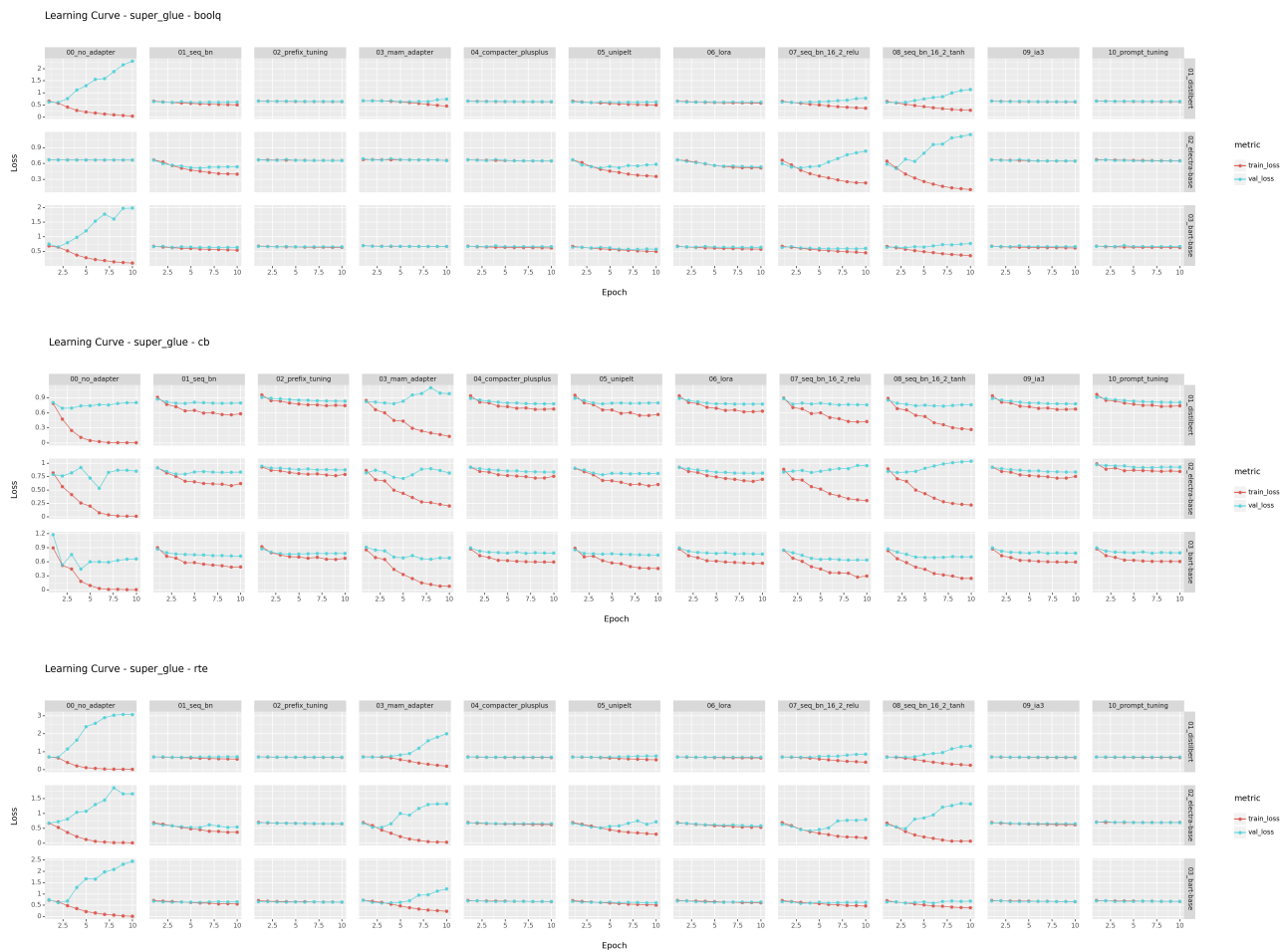

## D. Learning Curves by Task – GLUE

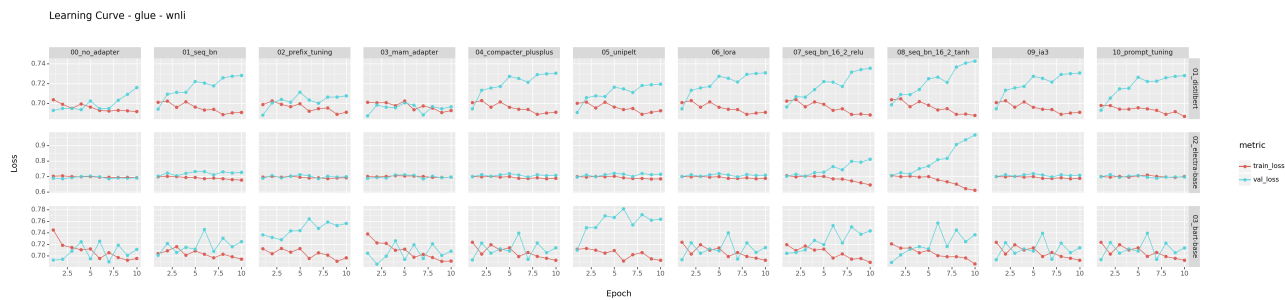

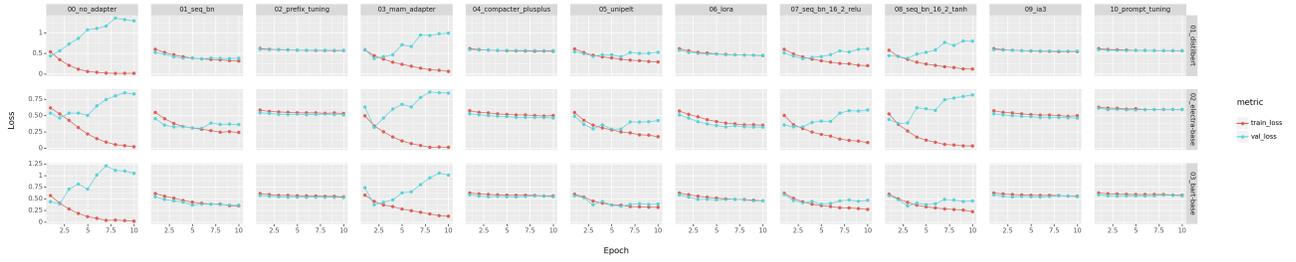

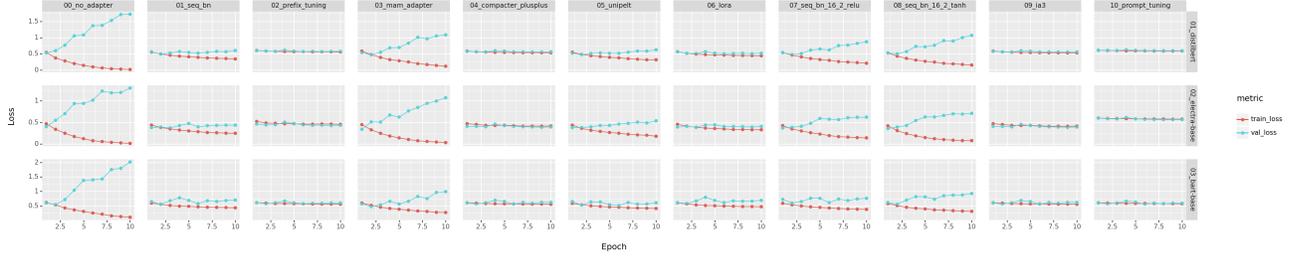

## E. Task Level Metrics – SuperGLUE

We also provide visualizations to compare validation and training set loss, accuracy, supplementary performance metrics like Matthew's Correlation, training and inference runtimes, and training time floating point operations by adapter and model for each individual task across SuperGLUE and GLUE benchmarks.

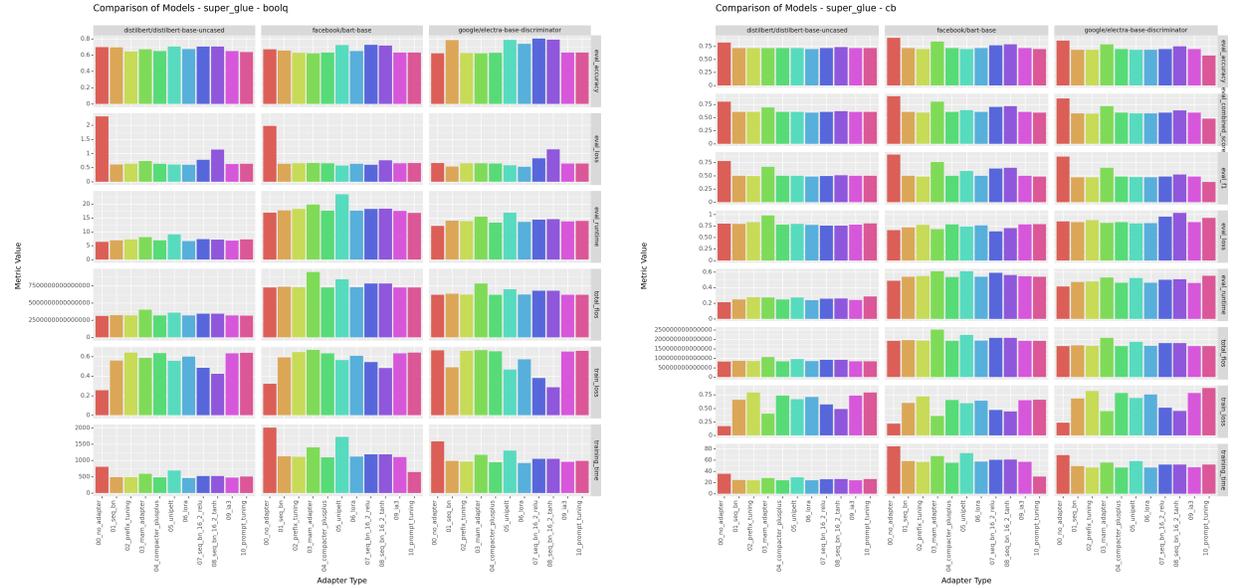

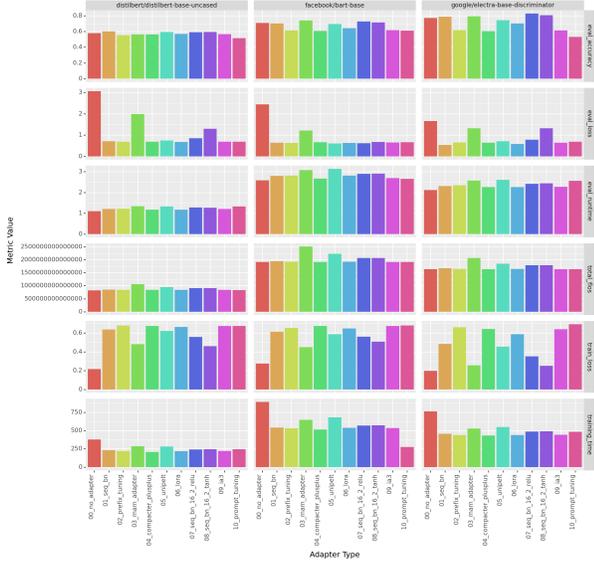
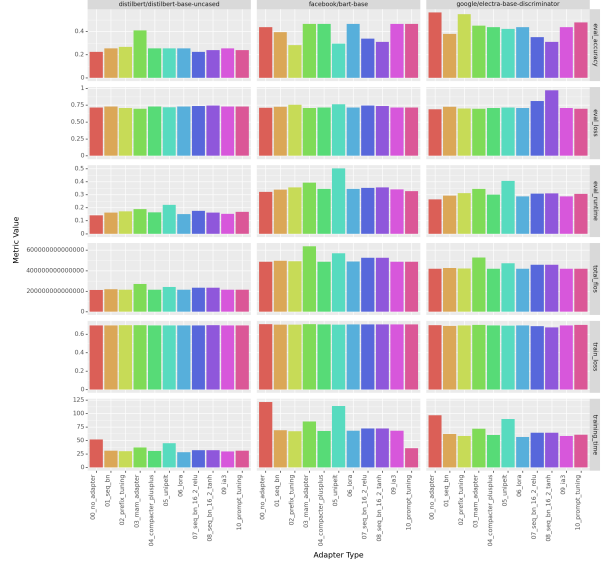
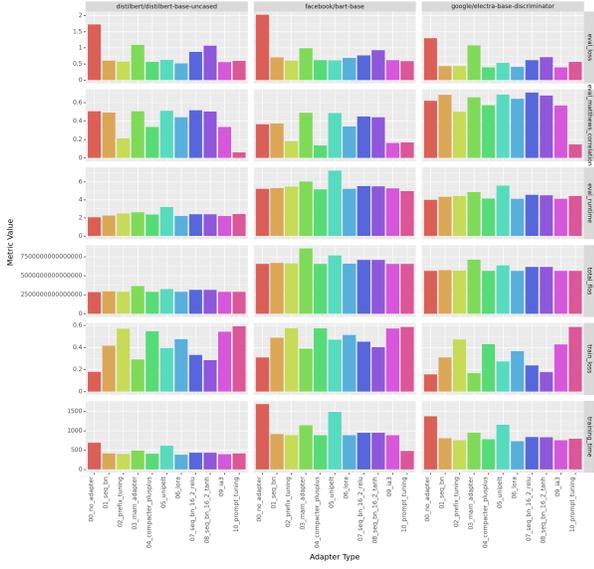
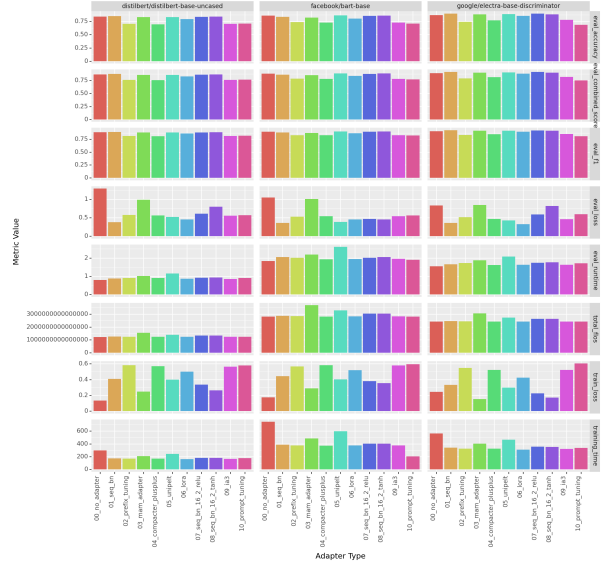

## F. Normalized Performance by Benchmark

We extend the visualization presented in section 4.1 to include the GLUE benchmark. Interestingly, a combination of DistilBERT and simpler adapters performs much better on GLUE than on SuperGLUE.

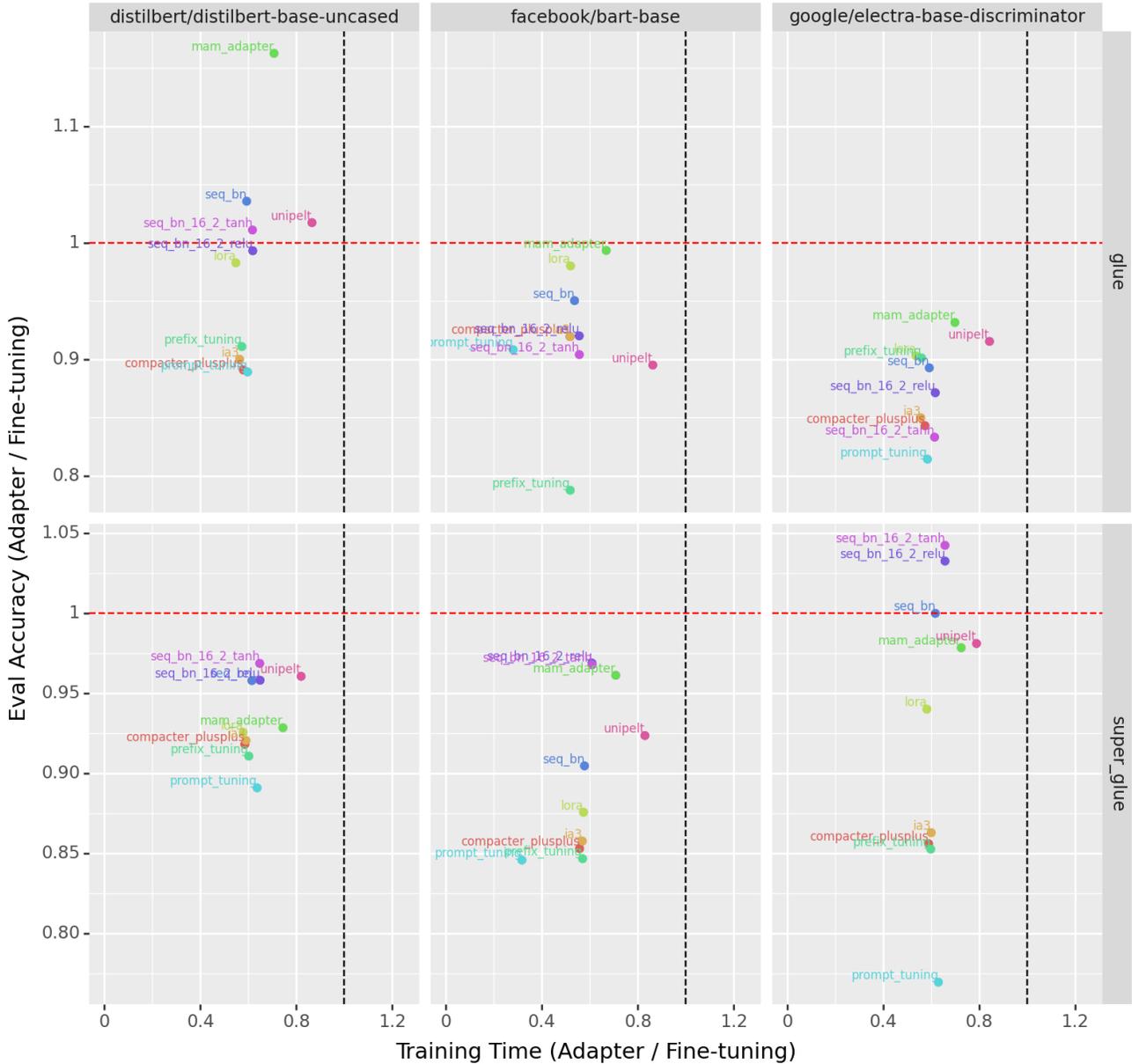

## G. News dataset – finetuning with different models

We initially aimed to experiment with news dataset with all models with different adapter models, before we reduced the scope to just Electra. Electra-small had least over fitting among all the models with finetuning.

| Model | Eval Loss | Eval Accuracy | Training Time |
|---|---|---|---|
| distilbert-base-uncased | 1.2752 | 0.6357 | 1:16:16 |
| facebook/bart-base | 1.1782 | 0.6588 | 2:51:55 |
| google/electra-base-discriminator | 1.3087 | 0.6391 | 2:33:41 |
| google/electra-small-discriminator | 1.7727 | 0.535 | 0:52:00 |

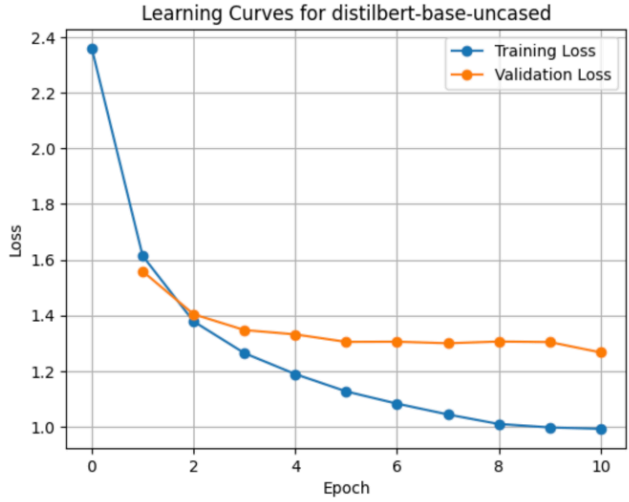

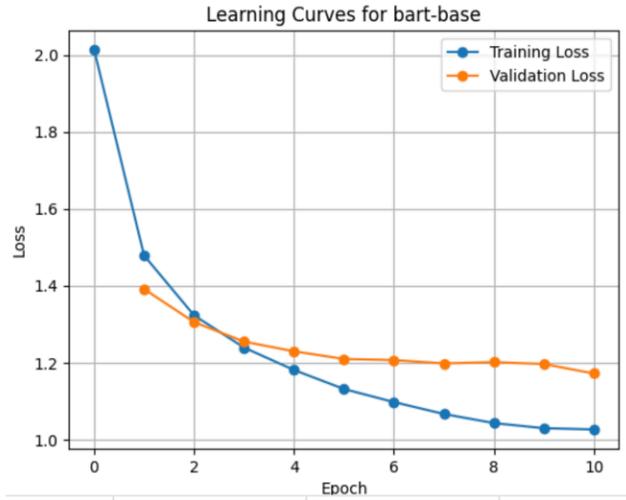

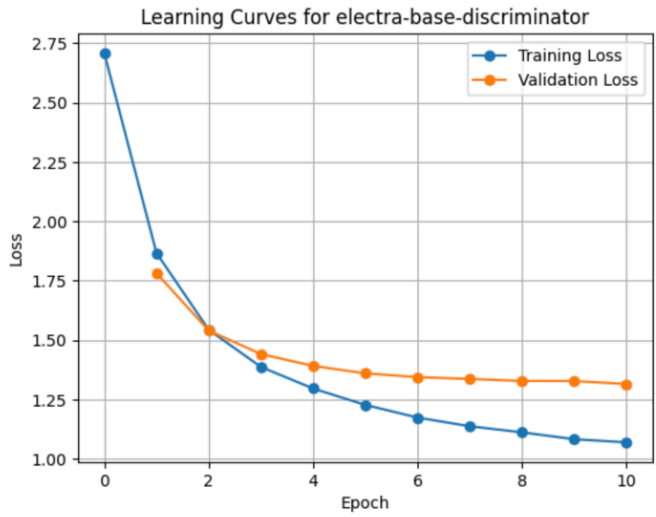

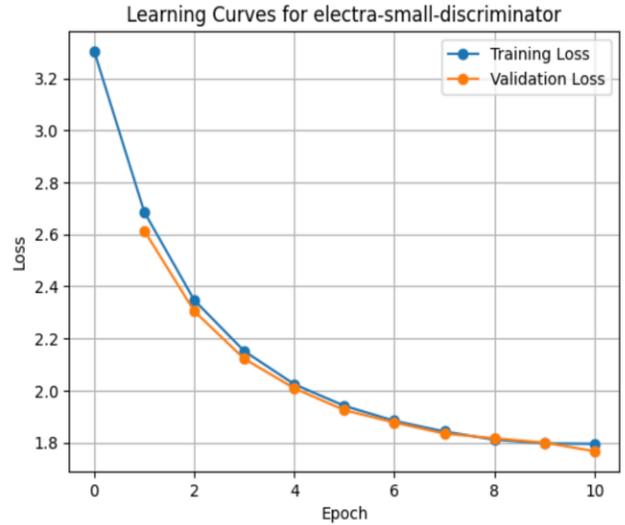